\documentclass[10pt,twocolumn,letterpaper]{article}

\usepackage{cvpr}              % To produce the CAMERA-READY version
\definecolor{cvprblue}{rgb}{0.21,0.49,0.74}
\usepackage[pagebackref,breaklinks,colorlinks,allcolors=cvprblue]{hyperref}

\def\paperID{*****} % *** Enter the Paper ID here
\def\confName{CVPR}
\def\confYear{2026}

\title{Multimodal Causality-Driven Representation Learning for Generalizable Medical Image Segmentation}

\author{
    Xusheng Liang$^{1,2*}$, 
    Lihua Zhou$^{3*}$,  
    Nianxin Li$^{4}$,
    Miao Xu$^{3,5,6}$,
    Ziyang Song$^{3}$,
    Dong Yi$^{3}$, \\
    Jinlin Wu$^{3,5\dagger}$,
    Jiawei Ma$^{1\dagger}$,
    Hongbin Liu$^{3,5}$,
    Zhen Lei$^{3,5,6}$,
    Jiebo Luo$^{3}$
     \\
    {\normalsize $^1$City University of Hong Kong \quad $^2$Shenzhen Loop Area Institute} \\
    {\normalsize $^3$CAIR, HKISI, Chinese Academy of Sciences}
    {\normalsize $^4$UESTC} \\
    {\normalsize $^5$MAIS, Institute of Automation, Chinese Academy of Sciences} \\
    {\normalsize $^6$School of Artificial Intelligence, University of Chinese Academy of Sciences} \\
    {\tt\small xushliang2-c@my.cityu.edu.hk, jinlin.wu@cair-cas.org.hk, jiaweima@cityu.edu.hk}  
}

\begin{document}
\maketitle

\begingroup
\renewcommand\thefootnote{}
% \footnotetext{*Equal contribution. \\ \ $^\dagger$Corresponding author.}
\footnotetext{%
\begin{tabular}{@{}r@{\ }l@{}}
$^*$ & Equal contribution. \\
$^\dagger$ & Corresponding authors.
\end{tabular}
}
\endgroup

\begin{abstract}
Vision-Language Models (VLMs), such as CLIP, have demonstrated remarkable zero-shot capabilities in various computer vision tasks. However, their application to medical imaging remains challenging due to the high variability and complexity of medical data. Specifically, medical images often exhibit significant domain shifts caused by various confounders, including equipment differences, procedure artifacts, and imaging modes, which can lead to poor generalization when models are applied to unseen domains. To address this limitation, we propose Multimodal Causal-Driven Representation Learning (MCDRL), a novel framework that integrates causal inference with the VLM to tackle domain generalization in medical image segmentation. MCDRL is implemented in two steps: first, it leverages CLIP's cross-modal capabilities to identify candidate lesion regions and construct a confounder dictionary through text prompts, specifically designed to represent domain-specific variations; second, it trains a causal intervention network that utilizes this dictionary to identify and eliminate the influence of these domain-specific variations while preserving the anatomical structural information critical for segmentation tasks. Extensive experiments demonstrate that MCDRL consistently outperforms competing methods, yielding superior segmentation accuracy and exhibiting robust generalizability.
\end{abstract}

\section{Introduction}

\begin{figure}
    \centering
    \includegraphics[width=0.95\linewidth]{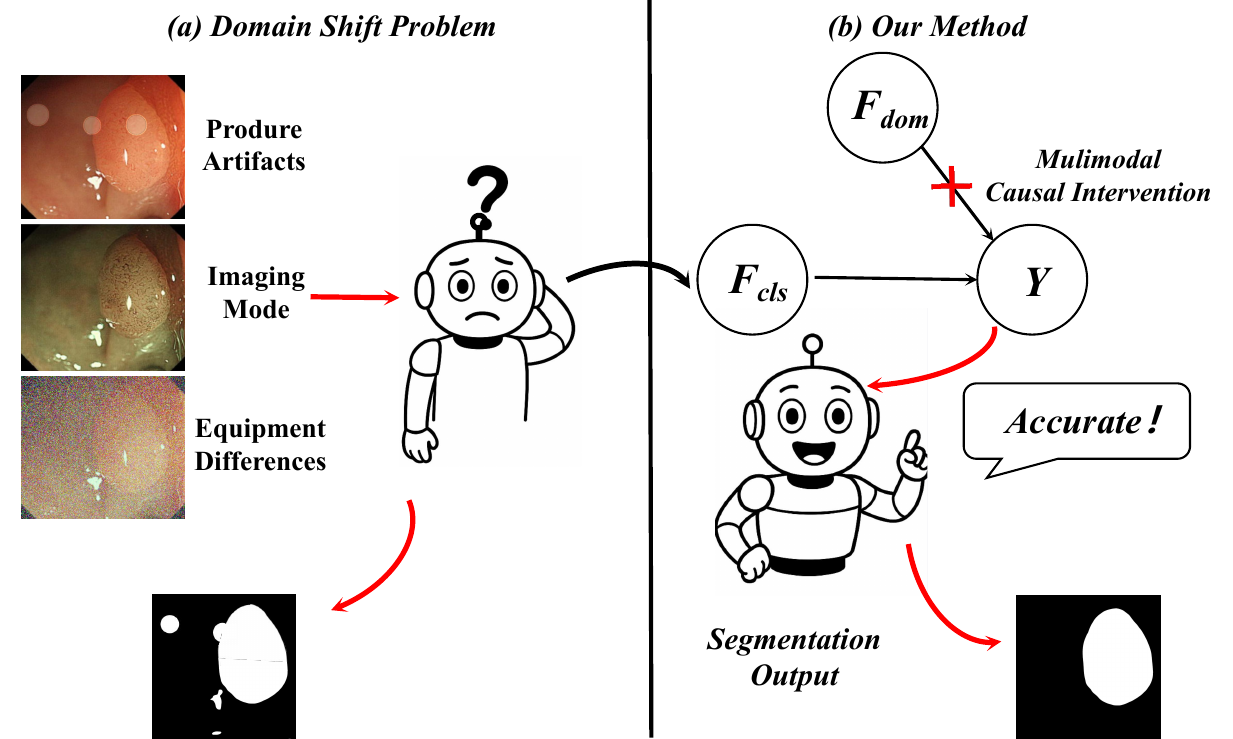}
    \caption{(a) Domain shift problems in medical images confuse the segmentation model, leading to inaccurate outputs. (b) Our method leverages multimodal causal intervention to remove confounders, resulting in more accurate and reliable outcomes.}
    \label{fig:1}
\end{figure}

Vision-Language Models (VLMs) have demonstrated impressive performance in various natural image tasks, leveraging their ability to align visual and textual representations through contrastive learning \cite{radford2021learning}. However, when applied to medical imaging, these models often struggle due to the inherent complexity and variability of medical data. Medical images exhibit significant domain shifts caused by differences in equipment, procedure artifacts, and imaging modes, leading to poor generalization across unseen scenarios \cite{karani2018lifelong, guan2021domain, kamnitsas2017unsupervised}. This limitation is particularly critical in specialized tasks such as endoscopic image segmentation, where accurate and robust performance across diverse domains is essential for clinical decision-making \cite{chamarthi2024mitigating, he2023domain}.

To address this limitation, Domain Generalization (DG) \cite{gulrajani2020search} has emerged as a promising direction for enhancing model adaptability in medical imaging. DG aims to equip models with the ability to generalize effectively to unseen domains by learning domain-invariant representations that are less sensitive to domain-specific variations. Existing DG methods for medical image segmentation typically employ various strategies to enhance model robustness across domains. These include domain adversarial training that uses discriminator networks to produce domain-agnostic representations \cite{li2018deep,ganin2016domain}, feature disentanglement techniques that separate anatomical features from domain-specific factors \cite{piratla2020efficient}, and meta-learning strategies that simulate domain shifts during training \cite{dou2019domain}. Despite these advances, they fail to explicitly address the removal of confounders, which are the key factors causing domain shift. As a result, these approaches are still unable to ensure accurate model outputs, as illustrated in Fig. \ref{fig:1}(a). 

% 尽管这些方法取得了一定的成功，但是这些方法并没有考虑如何显式的去除Confunder，它造成domian shift的关键因素，因此依旧无法使模型有着准确的输出，如图1左边所示。
% \subsection{Proposed Approach}

% 为了解决这个问题，我们引入了因果学习。如图1右边所示，我们通过构建了一个因果干预来去除confunder的影响，使得模型可以直接建模 类别固有信息(这个词可能要换一下)与类别之间的关系，从而提升了模型的泛化能力。
To address the challenges posed by domain shift, we propose Multimodal Causal-Driven Representation Learning (MCDRL), as illustrated in Fig. \ref{fig:1}(b). Our framework leverages causal intervention \cite{yao2021survey} to explicitly eliminate the influence of confounders, allowing the model to directly learn the intrinsic properties of each class and the relationships between them. By integrating causal inference with multimodal representation learning, MCDRL effectively bridges visual and semantic information, enabling robust and domain-invariant representation learning. This approach significantly enhances the model's generalization capabilities across diverse and unseen domains.

% 我们的方法可以大致分为两步，第一步，利用CLIP的跨模态能力，使用文本来构建了一个confunder字典。
% 第二步，基于这个confunder字典，我们训练了一个因果干预网络来对CLIP的视觉提取的特征进行因果干预，使其得到域不变特征，并进一步基于域不变特征训练一个分类器，直接建模类别固有信息(这个词可能要换一下)与类别之间的关系。我们的贡献总结为： 
% Our method can be summarized in two main steps. First, we utilize the cross-modal capabilities of CLIP to construct a confounder dictionary by generating style word vectors through text prompt. \zhou{These style vectors, combined with category information and segmentation pixel labels, are used to create text embeddings that simulate visual representations.}
% TODO: 这些text prompt 用作两个方面，一方面是为了与视觉embedding计算出可能的病变区域。另一方面也用做构建confunder字典。
Our method is implemented in two main steps. First, we utilize the cross-modal capabilities of CLIP to process text prompts in two ways. On one hand, the text prompts are used to compute potential lesion regions by aligning the visual embeddings with the textual embeddings. On the other hand, the text prompts are leveraged to construct a confounder dictionary, where diverse domain-related text embeddings are generated. Second, based on this confounder dictionary, we train a causal intervention network to perform causal interventions on the visual features extracted by CLIP. This process systematically eliminates spurious correlations introduced by domain-specific confounders, enabling the extraction of domain-invariant features to directly model the intrinsic properties of each category and the relationships between them. 

By integrating causal interventions with visual-language models, our method addresses the limitations of existing domain generalization methods and achieves superior performance in diverse clinical contexts, with a solid theoretical foundation grounded in causal inference.

Our work makes the following contributions:

\begin{itemize}
    \item We propose a novel method to integrate causal inference with visual-language models for medical image segmentation. By leveraging a multimodal confounder dictionary and causal interventions, our approach learns domain-invariant features with clear theoretical grounding, effectively addressing domain shift issues.
    \item We leverage CLIP's cross-modal capabilities to design a lesion region selection method and propose a causal intervention dictionary with supervised training. This approach effectively integrates causal learning into the VLM framework, enabling robust generalization while preserving multimodal alignment.
    \item We conduct extensive experiments on endoscopic datasets, demonstrating that our approach outperforms competitive methods and achieves robust generalization.

\end{itemize}

\section{Related Work}

\textbf{Vision Language Models.} Vision Language Models (VLMs) have seen rapid advancements driven by large-scale image-text datasets. CLIP \cite{radford2021learning} is the first to align image-text pairs in a shared embedding space using contrastive learning, demonstrating strong adaptability across domains. Subsequent works have improved VLMs through enhanced loss functions \cite{wu2024saco}, noise reduction in web-scale datasets \cite{zhou2022conditional, li2023blip,chen2025map}, and optimized training strategies \cite{sun2023eva}. These improvements have enabled VLMs to excel in tasks like zero-shot classification \cite{khattak2023maple, zhou2022conditional, li2026dvla, li2025vt,ma2024mode,yang2022tempclr}, domain generalization \cite{addepalli2024leveraging, laurenccon2024matters}, and dense prediction in 2D and 3D \cite{dong2023maskclip, peng2024parameter, qin2024langsplat}. With these advancements, VLMs now serve as powerful backbones for downstream tasks, reducing the need for task-specific training and offering robust solutions for domain-specific challenges.

{\textbf{Causal Inference for Robust Visual Recognition.} Causal inference is critical for robust visual recognition, as it helps uncover causal relationships and mitigate the effects of confounding factors \cite{pearl2000models}. For instance, models trained on airplane images may mistakenly associate airplanes with blue skies due to spurious correlations in the data \cite{arjovsky2019invariant}. To address this, diversifying the training contexts (e.g., incorporating images with varied backgrounds) enables models to focus on learning causal features rather than spurious ones. Structural causal models (SCMs) \cite{pearl2000models,zhou2025text} provide a formal framework for representing causal relationships through directed acyclic graphs (DAGs), with interventions denoted as \( do(\cdot) \), which simulate external manipulations of variables. Techniques such as invariant risk minimization (IRM) \cite{arjovsky2019invariant} and contrastive learning \cite{mahajan2021domain} leverage these causal frameworks. Specifically, IRM seeks invariance across training environments to identify causal features, while contrastive learning reduces spurious correlations by emphasizing differences between positive and negative samples. These approaches have shown improvements in domain robustness and generalization to unseen instances \cite{atzmon2020causal}.}

\textbf{Domain Generalization.} Domain Generalization (DG) aims to train models on diverse source domains to improve robustness in unseen environments. DG strategies typically fall into four categories: (1) Domain alignment techniques reduce inter-domain discrepancies with methods such as moment matching \cite{muandet2013domain}, contrastive loss \cite{motiian2017unified}, and adversarial learning \cite{li2018deep, shao2019multi}; (2) Meta-learning approaches optimize models on synthetic meta-train/meta-test splits to extract generalizable knowledge \cite{balaji2018metareg, li2018learning}; (3) Data augmentation methods use transformations and adversarial gradients to enhance robustness \cite{volpi2019addressing, shi2020towards}; and (4) Self-supervised learning exploits unlabeled data to learn versatile representations \cite{carlucci2019domain, wang2020learning}.

\section{Problem Analysis}

In traditional medical image analysis, the observed representation $F$ 
typically entangles both category-relevant information $F_c$ and 
domain-specific confounders $F_d$. A predictor \( H(\cdot) \) is then trained to model 
$P(Y \mid F)$. However, existing domain generalization (DG) studies have 
shown that the presence of $F_d$ introduces spurious correlations, 
preventing the model from generalizing to unseen domains.

From a causal perspective, the goal is to eliminate the influence of 
domain-specific confounders $F_d$ and recover the invariant relationship 
between $F_c$ and the pixel-wise segmentation label $Y$, i.e., modeling the interventional distribution 
$P(Y \mid \text{do}(F))$. Ideally, an intrinsic representation should 
remain invariant under diverse confounding factors. However, since $F_c$ is not directly observable, we instead approximate this intervention by operating on the learned representation $F$ and marginalizing over confounding factors. Specifically, a causal intervention module can be employed to perform backdoor adjustment via a global confounder dictionary. 

\begin{figure}[h!]
\centering
\includegraphics[width=1\linewidth]{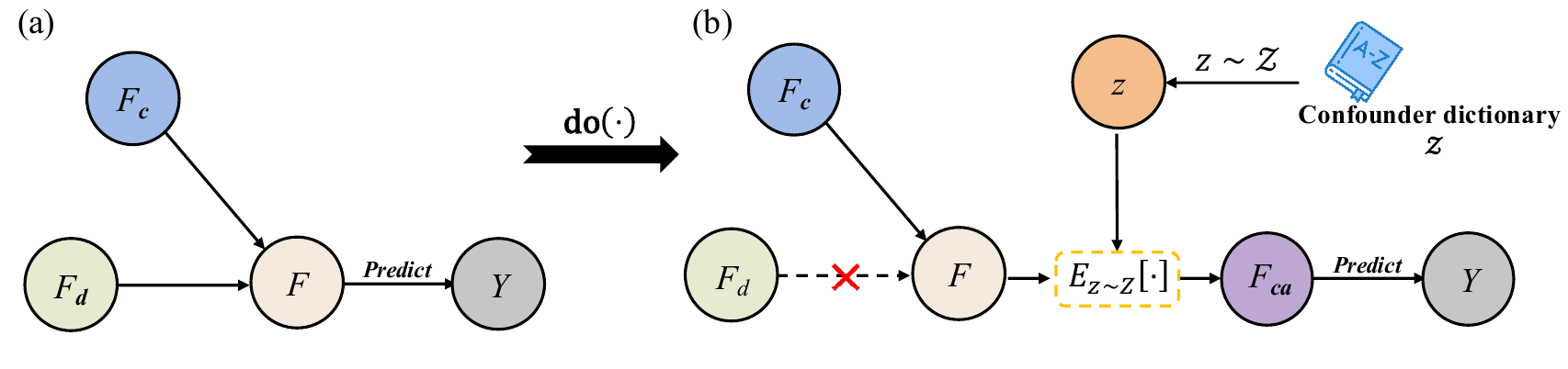}
% \vspace{-8pt}
\caption{
\textbf{Causal intervention via marginalization over confounders.}
(a) In the standard setting, the learned representation $F$ is influenced by both 
category-relevant features $F_c$ and domain-specific confounders $F_d$, leading to 
spurious correlations when predicting the label $Y$. 
(b) Our approach performs a causal intervention by blocking the direct effect of $F_d$ 
and marginalizing over its possible realizations. Specifically, we introduce a 
confounder dictionary $\mathcal{Z}$ and sample $z \sim \mathcal{Z}$ to account for 
the variability induced by confounders. The final representation is obtained via 
expectation $\mathbb{E}_{z \sim \mathcal{Z}}[\cdot]$, resulting in an invariant 
feature $F_{\text{inv}}$ that depends only on $F_c$. This yields an interventional 
distribution $P(Y \mid \text{do}(F))$, effectively suppressing spurious correlations 
induced by $F_d$.
}
\label{fig:scm}
\end{figure}

To approximate the interventional distribution $P(Y \mid \text{do}(F))$, 
we introduce a confounder dictionary $Z$ to model the distribution of 
domain-specific confounders. Instead of explicitly manipulating $F_d$, 
we approximate its effect by sampling $Z$ and marginalizing over confounding factors.

Following the framework of causal inference~\cite{pearl2000models}, 
the intervention can be expressed as:
\begin{equation}
P(Y \mid \text{do}(F)) = \sum_{z \in Z} P(Y \mid F, z) P(z),
\end{equation}
where $z$ represents a sampled confounder from the dictionary $Z$. 
This formulation integrates out the confounding factors, thereby 
eliminating the influence of $F_d$.

In practice, explicitly decomposing $F$ into $F_c$ and $F_d$ is infeasible. 
To address this, we introduce a causal intervention network $A(\cdot)$ 
that simulates the intervention process in feature space. Specifically, 
$A(\cdot)$ takes the observed representation $F$ and a sampled confounder $z$ 
as input, and outputs a transformed representation $F_z = A(F, z)$, 
which reflects the representation under a specific confounder.

The intervention process can then be reformulated as:
\begin{equation}
P(Y \mid \text{do}(F)) = \sum_{z \in Z} P(Y \mid A(F, z)) P(z).
\label{eq:backdoor}
\end{equation}

To make this computation tractable, we approximate the summation as an expectation:
\begin{equation}
P(Y \mid \text{do}(F)) = \mathbb{E}_{z} \left[ P(Y \mid A(F, z)) \right].
\end{equation}

We further introduce a segmentation head $H(\cdot)$ that maps the transformed 
features to predictions:
\begin{equation}
P(Y \mid A(F, z)) = H(A(F, z)).
\end{equation}

Thus, the intervention can be approximated as:
\begin{equation}
\begin{aligned}
P(Y \mid \text{do}(F)) 
&= \mathbb{E}_{z} \left[ H(A(F, z)) \right] \\
&\approx H \left( \mathbb{E}_{z} \left[ A(F, z) \right] \right),
\end{aligned}
\label{Eq:4}
\end{equation}
where the second approximation is introduced for computational efficiency.
Instead of averaging predictions over all confounder realizations,
we apply the prediction head to the averaged feature representation,
which serves as a tractable surrogate for the interventional inference.
The averaged representation $\mathbb{E}_{z}[A(F, z)]$ corresponds to an 
invariant feature $F_{\text{ca}}$, which is robust to variations in 
confounding factors. Based on Eq.~\eqref{Eq:4}, we design our MCDRL 
framework in the next section.

\section{Method}
\label{sec:method}

\noindent\textbf{Problem setting.} 
% Our multi-source domain generalization setting, 
Assume we have access to $N$ source domains $\{D_{s_1}, D_{s_2}, ..., D_{s_N}\}$ during training, where each source domain $D_{s_i} = \{(X^{s_i}_j, Y^{s_i}_j)\}^{n_{s_i}}_{j=1}$ consists of $n_{s_i}$ labeled samples with images $X^{s_i}_j \in \mathbb{R}^{H \times W \times 3}$ and their corresponding segmentation masks $Y^{s_i}_j \in \mathbb{R}^{H \times W \times K}$ for {$K$} distinct lesion classes $\{class_k\}_{k=1}^{K}$ (e.g., Polyps, Tumors, Inflam, Nodules, Cyst), where $ H $ and $ W $ are the spatial dimensions. Here, $K$ denotes the number of lesion classes, and
$Y^{s_i}_j$ is the corresponding pixel-wise annotation. Our objective is to train a model using these source domains that can generalize to an unseen target domain $D_t = \{X^t_j\}^{n_t}_{j=1}$ consisting of $n_t$ test medical images.

% \noindent\textbf{Overview}
\begin{figure*}[t]
\centering
\includegraphics[width=1\linewidth]{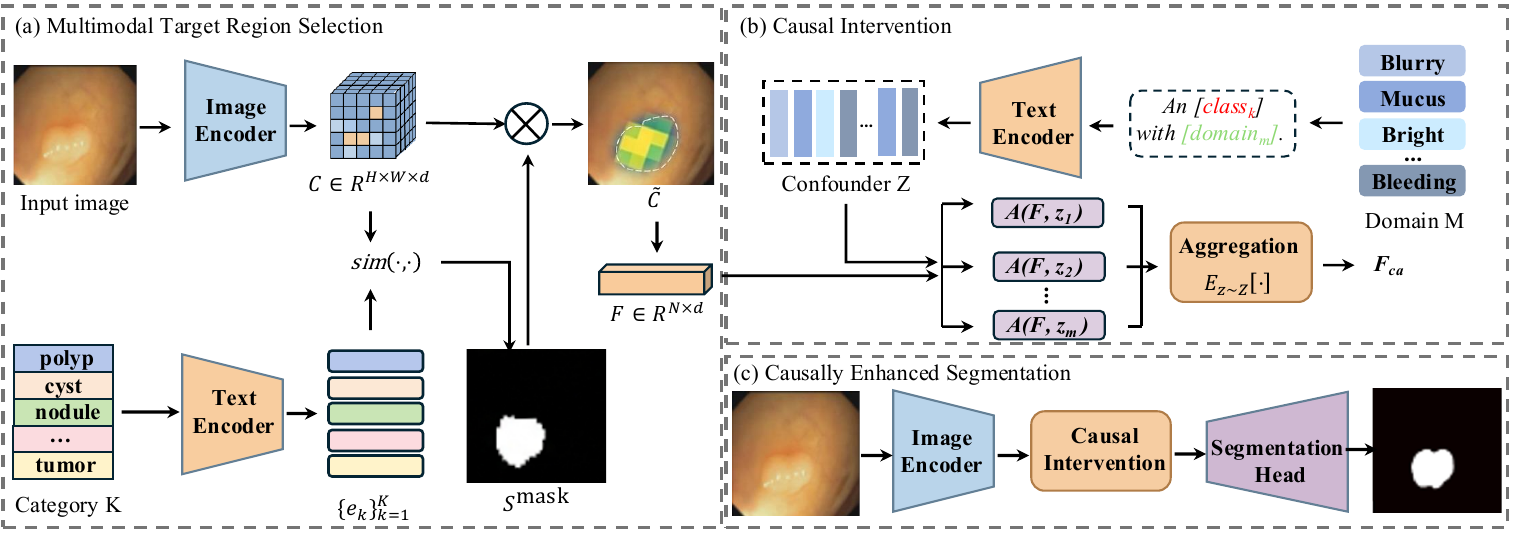}
\caption{Overview of the proposed MCDRL framework. 
(a) Multimodal Target Region Selection (MTRS) leverages text prompts to localize lesion-relevant regions via cross-modal similarity and extract region-aware features. 
(b) Causal-Driven Representation Learning (CDRL) models domain variations as confounders and performs causal intervention by marginalizing over a confounder dictionary, producing domain-invariant representations that suppress spurious correlations. 
(c) Causally-Enhanced Segmentation feeds the invariant features into the decoder to generate the final lesion mask.}
\label{fig:framework}
\end{figure*}

\noindent\textbf{Overview.}
We propose a multimodal causal intervention framework for lesion segmentation in medical images. As shown in Fig.~\ref{fig:framework}, our method consists of two innovative components: (1) \textbf{Multimodal Target Region Selection (MTRS)}, which leverages multimodal knowledge encoded in text prompts to highlight potential lesion regions, addressing the challenge of lesion localization. (2) \textbf{Causal-Driven Representation Learning (CDRL)}, which performs causal intervention to eliminate the influence of imaging condition confounders, ensuring the model learns from invariant pathological features. These components feed into a segmentation head that produces lesion masks.

\subsection{Multimodal Target Region Selection} 

To construct the observed representation $F$, we first extract visual features from the input image using CLIP's vision encoder. For an arbitrary source domain image $X^{s_i}_j$, the vision encoder generates a dense feature map $C \in \mathbb{R}^{H \times W \times d}$, where $d$ is the feature dimensionality. Each spatial location $(h, w)$ in $C$ corresponds to a local visual feature vector $C[h, w] \in \mathbb{R}^d$, capturing rich visual information.

Simultaneously, we generate a set of textual embeddings $\{e_k\}_{k=1}^K$ using CLIP's text encoder. These embeddings are derived from the template:  
\textit{``A \{class\textsubscript{k}\} in an endoscopic image''},  
where $class_k$ represents one of $K$ predefined categories relevant to the segmentation task. Each textual embedding $e_k \in \mathbb{R}^d$ encodes semantic information about the corresponding category. The resulting set of textual embeddings $\{e_k\}_{k=1}^K$ encapsulates high-level semantic priors, providing a robust reference for spatial alignment and selecting possible lesion regions.

To align the visual and textual features, we compute a similarity tensor $S \in \mathbb{R}^{H \times W \times K}$ that measures the correspondence between the visual feature map $C$ and the textual embeddings $\{e_k\}_{k=1}^K$. Specifically, for each spatial location $(h, w)$, we calculate the cosine similarity between the local visual feature $C[h, w]$ and each textual embedding $e_k$. The similarity score for category $k$ at location $[h, w]$ is given by:
\begin{equation}
S[h, w, k] = \frac{C[h, w] \cdot e_k}{\|C[h, w]\| \|e_k\|},
\end{equation}
where $\cdot$ denotes the dot product and $\|\cdot\|$ is the L2-norm. The resulting similarity tensor $S$ captures the alignment between each spatial location and all $K$ categories. 

To obtain a unified similarity map $\hat{S} \in \mathbb{R}^{H \times W}$, we aggregate the similarity scores across categories by selecting the maximum value at each spatial location, i.e.,
\[
\hat{S}[h, w] = \max_{k \in \{1, \cdots, K\}} S[h, w, k].
\]
This map effectively highlights the most relevant category for each spatial location.

Based on the unified similarity map $\hat{S}$, we generate a binary mask $S^{\mathrm{mask}} \in \{0, 1\}^{H \times W}$ to select the most relevant regions for the segmentation task. Specifically, we rank the similarity scores in $\hat{S}$ and retain the top $N = \alpha \cdot H \cdot W$ regions with the highest scores, where $\alpha \in [0, 1]$ is a hyperparameter controlling the sparsity of the mask. This mask isolates the most relevant regions in the image while suppressing irrelevant background areas.

Finally, we extract the region features by performing element-wise multiplication between the original feature map $C$ and the binary mask $S^{\mathrm{mask}}$. Specifically, for each spatial location $(h, w)$, we compute:
% \begin{equation}
$
\tilde{C}[h, w] = C[h, w] \, S^{\mathrm{mask}}[h, w],
% \end{equation}
$
We keep the non-zero entries in $\tilde{C}$ and reshape it into a compact representation $F \in \mathbb{R}^{N \times d}$, where $N$ is the number of selected regions. This feature $F$ captures both category-related and domain-specific information, serving as the observed representation.

\subsection{Causal-Driven Representation Learning.}

\subsubsection{Confounder Dictionary Initialization.}

To model domain-specific confounders, we construct a confounder dictionary $ Z $ using text prompts. 
Each prompt describes a typical domain variation in endoscopic imaging, e.g., \textit{``An endoscopy image with \{domain\textsubscript{n}\}.''} The confounders are organized into five clinically relevant categories, including: 
(i) view quality (e.g., blur, artifacts), 
(ii) illumination conditions (e.g., bright, dim, uneven lighting), 
(iii) imaging techniques (e.g., narrow-band imaging, white-light imaging), 
(iv) distance factors (e.g., close-up, distant view), and 
(v) surface interference (e.g., mucus, blood, reflection). 
These factors align with established clinical guidelines and have been refined by expert consultation, capturing the dominant sources of domain shift encountered in routine endoscopic practice.

We define $ M = 12 $ representative prompts from these categories. 
The choice of $ M $ is design-driven, balancing coverage of domain variations and computational efficiency. The prompts are encoded by the text encoder to obtain $ Z = \{z_m\}_{m=1}^{M} $, where each $ z_m \in \mathbb{R}^{d} $, and equivalently $ Z \in \mathbb{R}^{M \times d} $. This dictionary serves as a discrete approximation of the confounder distribution for causal modeling.

\subsubsection{Domain-Invariant Representation Learning.}

To reduce the impact of domain-specific biases in the visual features $ F $, we apply causal intervention based on the confounder dictionary. Specifically, we model confounders as a dictionary $ Z = \{z_1, z_2, \ldots, z_M\} $ and perform feature refinement by marginalizing over all confounder values, as described in Eq.~\eqref{Eq:4}.
\begin{equation}
 \text{do}(F) \approx \mathbb{E}_{z} \left[ A(F, z) \right] \approx A(F, Z).
\end{equation}
Since direct computation of this expectation is intractable, we approximate the marginalization using a cross-attention mechanism, which efficiently captures the interaction between the selected features $ F \in \mathbb{R}^{N \times d} $ and the confounder dictionary $ Z \in \mathbb{R}^{M \times d} $.

Therefore, the do-operation for each extracted image feature is formulated as:
\begin{equation}
F_{\mathrm{ca}} = A(F, Z) = \operatorname{Attn}(F, Z),
\end{equation}
where $ F_{\mathrm{ca}} \in \mathbb{R}^{N \times d} $ denotes the domain-invariant feature after causal intervention, and $ A(\cdot, \cdot) $ represents a cross-attention mechanism that models the relevance of each confounding factor in the dictionary $ Z $ to the input feature $ F $.

% Once the causally intervened feature $F_{\mathrm{ca}}$ is obtained, it is fed into a segmentation head to produce the final prediction map $P \in [0,1]^{H \times W \times K}$, which represents the pixel-wise class probabilities.

Once the causally intervened feature $F_{\mathrm{ca}} \in \mathbb{R}^{N \times d}$ is obtained, 
it is passed through a segmentation head $H(\cdot)$ to produce the final prediction map:
$P = H(F_{\mathrm{ca}})$, where $P \in [0,1]^{H \times W \times K}$ denotes the pixel-wise class probabilities, with $P(h,w,k)$ representing the predicted probability of class $k$ at spatial location $(h,w)$.

\subsubsection{Loss Functions.}

The total loss function is defined as:
\begin{equation}
\mathcal{L} = \mathcal{L}_{\text{seg}} + \lambda_1 \mathcal{L}_{\text{causal}} + \lambda_2 \mathcal{L}_{\text{contrast}},
\end{equation}

where $ \mathcal{L}_{\text{seg}} $ is the segmentation loss calculated using the cross-entropy loss to ensure pixel-wise classification:
\begin{equation}
\mathcal{L}_{\mathrm{seg}}
= - \sum_{h=1}^{H} \sum_{w=1}^{W} \sum_{k=1}^{K}
Y(h,w,k)\log P(h,w,k),
\end{equation}
where $ Y(h,w,k) $ denotes the label at spatial location $(h,w)$ for class $k$, and $ P(h,w,k) $ is the predicted probability.

The causal loss $ \mathcal{L}_{\text{causal}} $ measures the discrepancy between the intervened feature $ F_{\mathrm{ca}} $ and the average feature of each class $ k $ across all domain prompts $ z_m $, as follows:
\begin{equation}
\mathcal{L}_{\text{causal}} =
\left\| \bar{F}_{\mathrm{ca}} - \frac{1}{M} \sum_{m=1}^{M} t_{k,m} \right\|^2,
\end{equation}
where $ \bar{F}_{\mathrm{ca}} = \mathrm{Pool}(F_{\mathrm{ca}}) \in \mathbb{R}^{d} $ denotes a pooled representation of $F_{\mathrm{ca}}$, and $ t_{k,m} \in \mathbb{R}^{d} $ is the text embedding generated from the template \textit{``A [class\textsubscript{k}] with [domain\textsubscript{m}]''}.

Each input image $ X^{s_i}_j $ corresponds to a single disease type, and the disease category $ k $ is derived from the ground truth label $ Y^{s_i}_j $. To enhance the alignment between visual and textual features, we fine-tune CLIP's vision encoder using a contrastive loss $ \mathcal{L}_{\text{contrast}} $, defined as:
\begin{equation}
\mathcal{L}_{\text{contrast}} =
- \log \frac{\exp\left(\mathrm{sim}(e_k, F_{\mathrm{vis}}) / \tau\right)}
{\sum_{k'=1}^{K} \exp\left(\mathrm{sim}(e_{k'}, F_{\mathrm{vis}}) / \tau\right)},
\end{equation}
where $ e_k $ is the text embedding generated from the template \textit{``A \{class\textsubscript{k}\} in an endoscopic image''}, $ F_{\mathrm{vis}} \in \mathbb{R}^{d} $ is the image-level feature extracted by CLIP's vision encoder, and $\tau = 0.5$ controls the sharpness of the softmax distribution. This ensures the model learns discriminative visual features aligned with the corresponding disease class.

The weights $\lambda_1$ and $\lambda_2$ control the contributions of $\mathcal{L}_{\text{causal}}$ and $\mathcal{L}_{\text{contrast}}$, respectively, and are empirically set to $\lambda_1 = 0.5$ and $\lambda_2 = 0.1$ based on validation performance.

\section{Experiments}
\label{exp}

\subsection{Experiment setup}
\subsubsection{Dataset.}
We evaluate our method on five datasets spanning three distinct anatomical domains within the natural cavity, showcasing both multi-center and multi-anatomical characteristics. For bronchoscopy, we use the BM-BronchoLC dataset (Site A, \cite{vu2024bm}), which contains 3,057 annotated frames with pixel-level segmentation masks for various respiratory tract lesions. The laryngoscopy evaluation is conducted on the Laryngoscope8 dataset (Site B, \cite{yin2021laryngoscope8}), comprising 3,533 annotated images with segmentation masks for vocal tract pathologies. Additionally, we employ three laparoscopy segmentation datasets collected from different medical centers, including CVC-ClinicDB/CVC-612 (Site C, \cite{bernal2015wm}), which contains 612 images, ETIS (Site D, \cite{silva2014toward}) with 196 images, and Kvasir (Site E, \cite{jha2019kvasir}) with 1,000 images. 
\subsubsection{Metric.}
We evaluate model performance using three key metrics. The Dice Coefficient (Dice) measures the overlap between predicted and ground truth regions. The Intersection over Union (IoU) quantifies the ratio of the intersection to the union of predicted and ground truth regions. The Accuracy (Acc) assesses overall correctness.

\begin{table*}[htbp!]
\centering
\caption{\textbf{Multi-Source to Single Source Domain Generalization Results.} Comparison of different methods across multiple datasets using ResNet-50, ViT-B/16, and ViT-L/14 backbones. ``Site A" means training on Sites B-E and testing on Site A, and similarly for the others. The best performances are in \textbf{bold}.}
% \vspace{-8pt}
\label{tab:domain_generalization}
\renewcommand{\arraystretch}{1.2}
\setlength{\tabcolsep}{4pt}
\small
\begin{tabular}{l|ccc|ccc|ccc|ccc|ccc|ccc}
\toprule
\multirow{2}{*}{\textbf{Method}} & \multicolumn{3}{c|}{\textbf{Site A}} & \multicolumn{3}{c|}{\textbf{Site B}} & \multicolumn{3}{c|}{\textbf{Site C}} & \multicolumn{3}{c|}{\textbf{Site D}} & \multicolumn{3}{c|}{\textbf{Site E}} & \multicolumn{3}{c}{\textbf{Average}} \\
\cline{2-19}
& \textbf{Dice} & \textbf{IoU} & \textbf{Acc} & \textbf{Dice} & \textbf{IoU} & \textbf{Acc} & \textbf{Dice} & \textbf{IoU} & \textbf{Acc} & \textbf{Dice} & \textbf{IoU} & \textbf{Acc} & \textbf{Dice} & \textbf{IoU} & \textbf{Acc} & \textbf{Dice} & \textbf{IoU} & \textbf{Acc} \\
\midrule
\multicolumn{19}{c}{\emph{ResNet-50 with pre-trained weights from CLIP.}} \\
\midrule
Baseline  & 75.3 & 68.2 & 91.2 & 72.1 & 65.3 & 88.4 & 73.2 & 67.1 & 89.7 & 69.5 & 63.1 & 87.2 & 70.4 & 64.2 & 88.1 & 72.1 & 65.6 & 88.9 \\
StyLIP   & 78.1 & 71.1 & 92.5 & 75.4 & 68.2 & 90.2 & 76.3 & 69.5 & 91.5 & 73.2 & 66.8 & 89.3 & 74.1 & 67.5 & 90.1 & 75.4 & 68.6 & 90.7 \\
BiomedCoOp   & 79.2 & 72.6 & 93.1 & 76.6 & 69.7 & 90.8 & 77.5 & 70.8 & 92.1 & 74.8 & 68.4 & 90.1 & 75.6 & 69.1 & 90.9 & 76.7 & 70.1 & 91.4 \\
MCDRL      & \textbf{81.5} & \textbf{74.2} & \textbf{94.3} & \textbf{78.5} & \textbf{71.3} & \textbf{92.1} & \textbf{79.4} & \textbf{72.3} & \textbf{93.2} & \textbf{76.4} & \textbf{69.5} & \textbf{91.4} & \textbf{77.3} & \textbf{70.4} & \textbf{92.0} & \textbf{78.6} & \textbf{71.5} & \textbf{92.6} \\
\midrule
\multicolumn{19}{c}{\emph{ViT-B/16 with pre-trained weights from CLIP.}} \\
\midrule
Baseline  & 76.4 & 69.3 & 92.1 & 73.5 & 66.7 & 89.5 & 74.7 & 68.1 & 90.8 & 71.2 & 64.9 & 88.4 & 72.1 & 65.6 & 89.3 & 73.6 & 66.9 & 90.0 \\
StyLIP    & 79.1 & 72.2 & 93.2 & 76.8 & 69.7 & 91.3 & 77.8 & 70.8 & 92.4 & 74.5 & 67.9 & 90.2 & 75.6 & 68.8 & 91.1 & 76.8 & 69.9 & 91.6 \\
BiomedCoOp   & 80.3 & 73.4 & 93.8 & 77.9 & 70.8 & 91.9 & 78.8 & 71.9 & 93.0 & 76.1 & 69.3 & 91.1 & 76.8 & 70.1 & 91.8 & 78.0 & 71.1 & 92.3 \\
MCDRL      & \textbf{82.6} & \textbf{75.4} & \textbf{95.1} & \textbf{79.9} & \textbf{72.6} & \textbf{93.0} & \textbf{80.9} & \textbf{73.6} & \textbf{94.1} & \textbf{77.9} & \textbf{70.9} & \textbf{92.3} & \textbf{78.8} & \textbf{71.8} & \textbf{92.9} & \textbf{80.0} & \textbf{72.9} & \textbf{93.5} \\
\midrule
\multicolumn{19}{c}{\emph{ViT-L/14 with pre-trained weights from CLIP.}} \\
\midrule
Baseline  & 78.1 & 70.7 & 93.0 & 75.2 & 68.2 & 90.4 & 76.1 & 69.4 & 91.6 & 72.8 & 66.2 & 89.5 & 73.5 & 67.1 & 90.2 & 75.1 & 68.3 & 90.9 \\
StyLIP    & 80.6 & 73.5 & 94.1 & 78.2 & 70.9 & 92.1 & 79.2 & 72.3 & 93.2 & 76.1 & 69.3 & 91.3 & 77.2 & 70.1 & 92.0 & 78.3 & 71.2 & 92.5 \\
BiomedCoOp   & 81.7 & 74.6 & 94.6 & 79.3 & 72.1 & 92.7 & 80.4 & 73.4 & 93.8 & 77.5 & 70.5 & 92.0 & 78.3 & 71.4 & 92.7 & 79.4 & 72.4 & 93.2 \\
MCDRL      & \textbf{84.1} & \textbf{76.8} & \textbf{95.9} & \textbf{81.4} & \textbf{73.9} & \textbf{93.8} & \textbf{82.5} & \textbf{75.1} & \textbf{94.8} & \textbf{79.6} & \textbf{72.4} & \textbf{93.1} & \textbf{80.5} & \textbf{73.3} & \textbf{93.7} & \textbf{81.6} & \textbf{74.3} & \textbf{94.3} \\
\bottomrule
\end{tabular}
\end{table*}

\subsubsection{Implementation Details.}
The model was implemented in PyTorch on one NVIDIA A800 GPU and trained using $224 \times 224$ pixel input resolution for the pre-trained CLIP models. We employed the AdamW optimizer with an initial learning rate of 0.005. We adopted a progressive training strategy where the causal intervention mechanism was activated after the tenth epoch, with training continuing for a total of 50 epochs.

\subsection{Comparison}
\subsubsection{Quantitative Results.}

Table~\ref{tab:domain_generalization} summarizes the evaluation of our method against state-of-the-art approaches across five clinical sites. Our method consistently outperforms all baselines across all metrics and datasets. Our approach achieves superior performance across all backbone architectures and test sites. Using the ViT-L/14 backbone, our method achieves an average mDice of 81.6\%, representing an 6.5\% improvement over the baseline and 2.0\% over the strongest competitor, BiomedCoOp. Larger backbones yield significant gains, with average mDice increasing by 3.8\% from ResNet-50 to ViT-L/14, underscoring the importance of robust feature extraction for domain generalization. Performance varies across sites, with Site A achieving the highest mDice and Site D the lowest, reflecting differences in clinical data distribution. Our method demonstrates stronger generalization across clinical sites compared to others. 

% While StyLIP and BiomedCoOp improve over the baseline using style transfer and prompt learning, they fall short of our performance. This highlights the effectiveness of modeling and disentangling confounding factors through causal intervention for cross-domain lesion segmentation.

\begin{figure*}[t]
\centering
\includegraphics[width=0.90\linewidth]{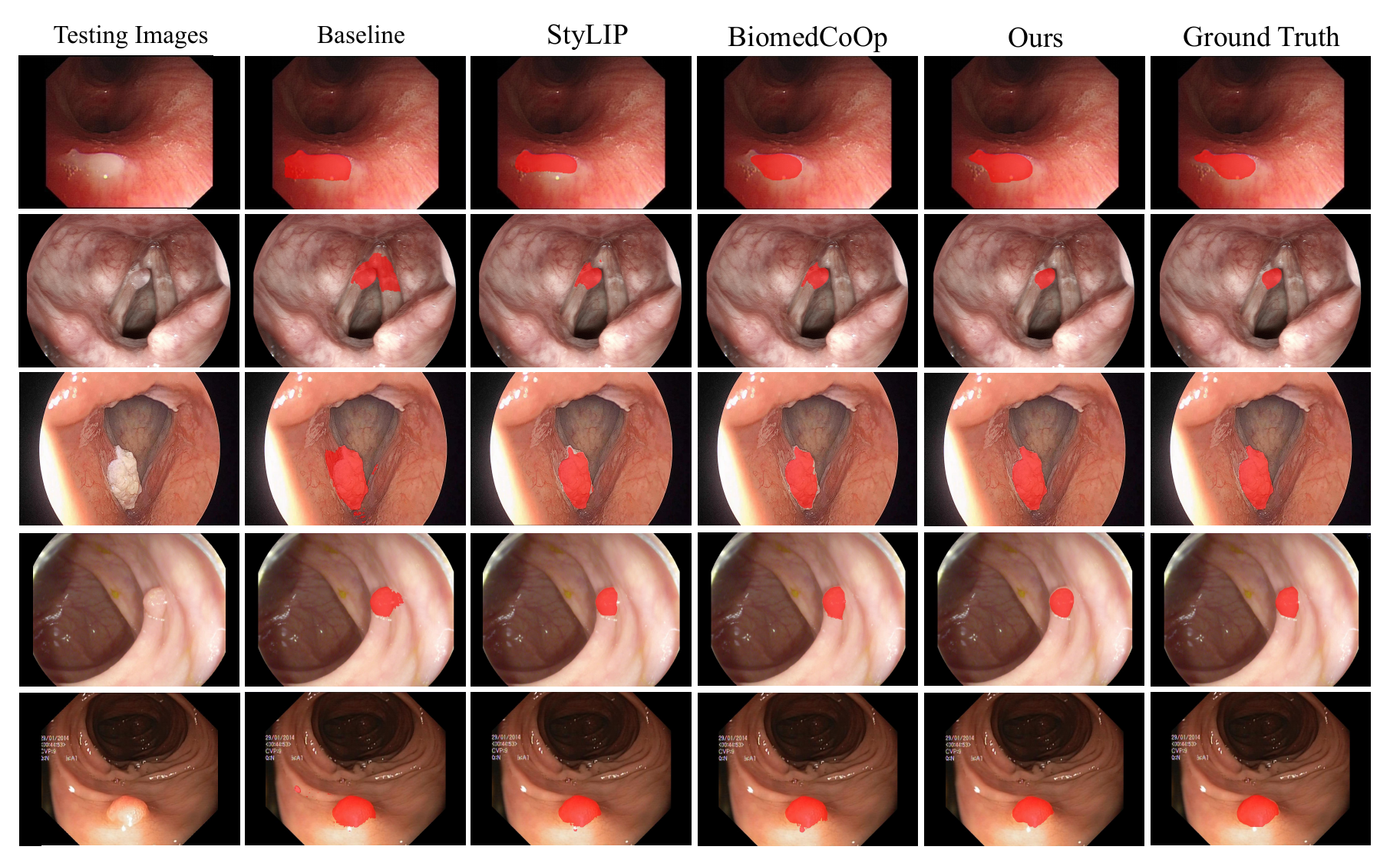}
\caption{Visual comparison of the segmentation results by the Baseline, StyLIP, BiomedCoOp, and our method in five datasets. The five rows from top to bottom display the final segmentation results for tests conducted on Sites A to E.}
\label{fig:comparison}
\end{figure*}

\begin{table}[htbp!]
\centering
\scriptsize
\begin{tabular}{@{}lcccccc@{}}
\toprule
\multirow{2}{*}{\textbf{Method}} & \textbf{Polyps} & \textbf{Tumors} & \textbf{Inflam.} & \textbf{Nodules} & \textbf{Cyst} & \textbf{Avg} \\
\cmidrule(lr){2-2} \cmidrule(lr){3-3} \cmidrule(lr){4-4} \cmidrule(lr){5-5} \cmidrule(lr){6-6} \cmidrule(lr){7-7}
 & mDice & mDice & mDice & mDice & mDice & mDice \\
\midrule
Baseline & 78.2 & 76.5 & 74.4 & 73.5 & 75.6 & 75.6 \\
StyLIP & 79.6 & 78.1 & 75.8 & 77.2 & 78.2 & 77.8 \\
BiomedCoOp & 81.5 & 79.8 & 77.3 & 78.5 & 75.6 & 78.5 \\
\textbf{MCDRL} & \textbf{83.9} & \textbf{82.2} & \textbf{80.1} & \textbf{84.9} & \textbf{80.8} & \textbf{82.4} \\
\bottomrule
\end{tabular}
\caption{Performance on different lesion types. The best performances are in \textbf{bold}.}
\label{tab:lesion_types}
% \vspace{-8pt}
\end{table}

Table~\ref{tab:lesion_types} demonstrates MCDRL's versatility across different lesion types. Our method shows consistent improvements over the baseline for all lesion categories: polyps (5.7\%), tumors (5.7\%), inflammatory lesions (5.7\%), nodules (11.4\%), and cysts (5.2\%). The most substantial improvement is observed for nodules (11.4\%), likely due to our method's ability to capture subtle textural and boundary characteristics critical for nodule identification.

The strong performance on inflammatory lesions is particularly significant as these typically present more subtle visual patterns that are often confounded by domain-specific acquisition factors. Similarly, the improvement in cyst segmentation demonstrates our method's effectiveness in handling lesions with variable appearances across different imaging protocols. Compared to the BiomedCoOp, MCDRL achieves an average improvement of 2.7\% across all lesion types. This consistent enhancement across diverse pathological conditions validates our approach's robustness and generalizability in real clinical applications.

As illustrated in Figure~\ref{fig:comparison}, our method achieves superior visual segmentation performance across diverse clinical scenarios from five different sites. The qualitative results demonstrate that MCDRL consistently produces more precise delineations with accurate boundary detection compared to baseline, StyLIP, and BiomedCoOp. While the baseline method often generates over-segmentation that extends beyond actual boundaries, both StyLIP and BiomedCoOp show progressive improvements but still exhibit imprecision. In contrast, our approach delivers segmentations that closely align with ground truth masks.

% % \vspace{-8pt}
\subsection{Further Studies}
% % \vspace{-8pt}
\begin{table}[htbp!]
\centering
\setlength{\tabcolsep}{3pt}
\small
\begin{tabular}{l|ccccc|c}
\hline
\multirow{2}{*}{\scriptsize Methods} & Site A & Site B & Site C & Site D & Site E & Avg. \\
\cline{2-7}
 & \multicolumn{6}{c}{\scriptsize Mean Dice(mDice) } \\
\hline\hline
Baseline & 65.60 & 74.65 & 63.15 & 68.11 & 75.34 & 69.37 \\
w/o MTRS & 77.09 & 83.24 & 80.50 & 78.32 & 83.18 & 80.47 \\
w/o CDRL & 79.51 & 79.36 & 77.40 & 77.29 & 80.00 & 78.71 \\
\hline
MCDRL & \textbf{82.53} & \textbf{88.98} & \textbf{88.73} & \textbf{90.25} & \textbf{91.83} & \textbf{88.46} \\
\hline
\end{tabular}
\caption{\small Ablation studies on different proposed modules. The best performances are in \textbf{bold}. ``Site A" means training on Sites B-E and testing on Site A, and similarly for the others.}
\label{tab:performance}
\end{table}

Table~\ref{tab:performance} presents an ablation study validating the contribution of each key component in our proposed approach. The baseline achieves an average mDice of 69.37\% across all five clinical sites. When removing the Multimodal Target Region Selection (MTRS) component while retaining the Domain-invariant Representation Learning (CDRL) mechanism, performance improves to 80.47\% mDice, demonstrating that causal modeling alone significantly enhances segmentation quality. Similarly, when eliminating CDRL while keeping MTRS, the model achieves 78.71\% mDice, indicating that region enhancement provides substantial benefits for anatomical structure identification.

Our full method, integrating both components, achieves the best performance with 88.46\%, showing a 19.09\% improvement over the baseline. This substantial gain highlights the complementary nature of these components: CDRL effectively disentangles domain-specific confounding factors from pathology-specific features, while MTRS focuses the model's attention on anatomically relevant regions. The performance improvement is particularly pronounced for challenging sites like Site C (25.58\%) and Site D (22.14\%), which present more complex visual characteristics that benefit most from our integrated approach.

% \vspace{-2pt}
\subsubsection{Ablation Study on Confounder Dictionary}
% \vspace{-8pt}
\begin{table}[ht]
\centering
\setlength{\tabcolsep}{12pt}
\small
\caption{Ablation study on the number of confounders $N$ in $Z$ using \textit{ViT-L/14}. Dice (\%) is reported across all datasets.}
% \vspace{-8pt}
\label{tab:ablation_study}
\begin{tabular}{@{}lccccc@{}}
\toprule
$N$               & 3    & 6    & 9    & 12   & 15   \\ \midrule
Site A              & 81.5 & 82.7 & 83.5 & 84.1 & 84.0 \\
Site B              & 78.9 & 79.6 & 80.5 & 81.4 & 81.2 \\
Site C              & 77.4 & 78.2 & 81.0 & 82.5 & 82.3 \\
Site D              & 75.6 & 77.3 & 78.4 & 79.6 & 79.4 \\
Site E              & 70.2 & 75.4 & 78.3 & 80.5 & 80.3 \\ \midrule
Avg.                & 76.7 & 78.6 & 80.3 & 81.6 & 81.4 \\ \bottomrule
\end{tabular}
\end{table}

% \vspace{-8pt}
Table~\ref{tab:ablation_study} evaluates the effect of the number of confounders $N$ in $Z$ on segmentation performance. Increasing $N$ generally enhances Dice scores across all sites, reflecting improved representation learning. As $N$ grows from 3 to 12, the average Dice rises from 76.7\% to 81.6\%, with Site A improving from 81.5\% to 84.1\%. However, beyond $N = 12$, the performance gains begin to plateau, with some sites showing slight decreases in Dice scores. For instance, for Site A, the Dice score drops marginally from 84.1\% at $N = 12$ to 84.0\% at $N = 15$, and a similar trend is observed for other sites. This suggests that excessive confounders may introduce redundancy or noise, which undermines the benefits of increased representational capacity. The consistency of this trend across all sites highlights the robustness and suggests that $N = 12$ provides an optimal balance of model complexity and performance.

% These results underscore the importance of carefully selecting the number of confounders to maximize the model's effectiveness without introducing unnecessary complexity.

\subsubsection{Ablation Study on Causal Network Depth}
% \vspace{-8pt}
\begin{table}[htbp]
\centering
\caption{ Impact of Causal Intervention Network depth on performance (Dice Score \%).}
% \vspace{-8pt}
\label{tab:causal_layers_ablation}
\resizebox{\columnwidth}{!}{ % 表格缩放到单栏宽度
\begin{tabular}{c|ccccc|c|cc}
\hline
\multirow{2}{*}{\# Layers} & \multicolumn{5}{c|}{Dice Score (\%)} & \multirow{2}{*}{Average} & \multirow{2}{*}{\begin{tabular}[c]{@{}c@{}}Params\\(M)\end{tabular}} & \multirow{2}{*}{\begin{tabular}[c]{@{}c@{}}Inference\\Time (ms)\end{tabular}} \\
\cline{2-6}
 & Site A & Site B & Site C & Site D & Site E & & & \\
\hline
1 & 81.3 & 78.7 & 79.2 & 76.4 & 77.8 & 78.7 & 12.4 & 142 \\
2 & 83.2 & 80.5 & 81.3 & 78.2 & 79.4 & 80.5 & 18.6 & 178 \\
3 & 84.1 & 81.4 & 82.5 & 79.6 & 80.5 & 81.6 & 24.9 & 215 \\
4 & 84.4 & 81.7 & 82.9 & 79.8 & 80.8 & 81.9 & 31.2 & 252 \\
5 & 84.5 & 81.8 & 83.0 & 80.0 & 80.9 & 82.0 & 37.5 & 290 \\
\hline
\end{tabular}
}
\end{table}

% \vspace{-10pt}
Table~\ref{tab:causal_layers_ablation} analyzes the influence of network depth on performance and efficiency. Deeper causal networks improve Dice scores across all sites, with the average rising from 78.7\% (1 layer) to 82.0\% (5 layers). Notably, Sites A and C gain over 3\% improvement, confirming the benefit of multi-layer causal reasoning. However, increasing depth also inflates computation: parameters grow from 12.4M to 37.5M and inference time from 142 ms to 290 ms. Performance saturates beyond 3--4 layers, suggesting that moderate depth achieves the best trade-off between accuracy and efficiency.

\subsubsection{Visual Evidence: Feature Distribution}

% \vspace{-1.2em}
\begin{figure}[h]
  \centering
  \includegraphics[width=0.85\linewidth]{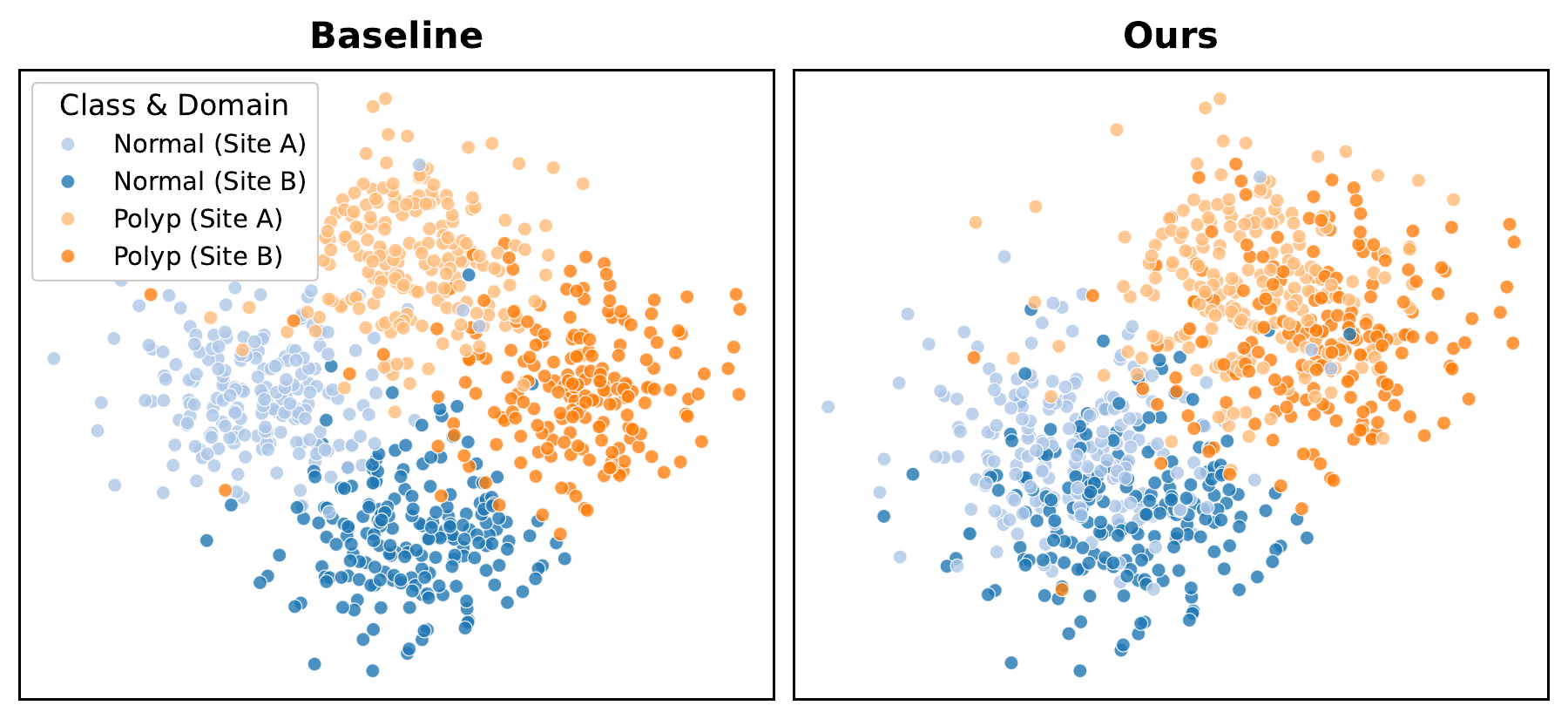}
  \vspace{-1.2em}
  \caption{\footnotesize \textbf{Class-Conditional t-SNE visualization.}}
  \label{fig:tsne_ext}
  % \vspace{-1.9em}
\end{figure}
\vspace{0.5em}

As shown in Fig.~\ref{fig:tsne_ext}, the class-conditional t-SNE visualization provides qualitative evidence of the effectiveness of our approach. The baseline exhibits fragmented intra-class clusters under domain shifts, reflecting sensitivity to acquisition variations, while MCDRL produces more compact and well-aligned class distributions across domains, indicating better capture of class-relevant semantics.

\section{Conclusion}

In this paper, we introduced Multimodal Causality-Driven Representation Learning (MCDRL), a framework that addresses domain generalization challenges in medical image segmentation by integrating causal inference with Vision-Language Models. Our approach leverages CLIP to identify candidate lesion regions and construct a confounder dictionary through text prompts, followed by a causal intervention network that eliminates domain-specific variations while preserving anatomical information. Extensive experiments across multiple datasets demonstrated MCDRL outperforms existing state-of-the-art methods in both segmentation accuracy and cross-domain robustness.

\clearpage
\section*{Acknowledgements}
This work was supported in part by NSFC under Grant 62502405 and the InnoHK program. We thank all our collaborators for their valuable discussions and support.

{
    \small
    \bibliographystyle{ieeenat_fullname}
    \bibliography{main}
}

% WARNING: do not forget to delete the supplementary pages from your submission 
% \input{sec/X_suppl}

\end{document}